\documentclass[runningheads]{llncs}
\usepackage{graphicx}
\usepackage{amsmath,amssymb} 
\usepackage{color}
\usepackage{float}
\usepackage{bbm}
\usepackage{soul}
\newcommand{\bd}[1]{\textbf{#1}}
\newcommand{\RR}{\mathbb{R}}

\newcommand{\printfnsymbol}[1]{%
	\textsuperscript{$\star \star$}%
}

\begin{document}
	\pagestyle{headings}
	\mainmatter
	\def\ECCV18SubNumber{2800}  
	
	\title{Improving Semantic Segmentation via Decoupled Body and Edge Supervision} 
	
	\titlerunning{Decoupled Body and Edge Segmentation}
	\authorrunning{X. Li et al.}
	
	\author{Xiangtai Li \inst{1} \thanks{ \scriptsize{Work done while at SenseTime. Email: lxtpku@pku.edu.cn}} \and  
		Xia Li \inst{1,2} \and 
		Li Zhang \inst{3} \and 
		Guangliang Cheng \inst{4} \thanks{\scriptsize{Correspond to: chengguangliang@sensetime.com,~yhtong@pku.edu.cn,~lz@robots.ox.ac.uk}} \and 
		Jianping Shi \inst{4} \and 
		Zhouchen Lin \inst{1} \and 
		Shaohua Tan \inst{1} \and 
		Yunhai Tong \inst{1} \printfnsymbol{1} 
	}

	\institute{Key Laboratory of Machine Perception, MOE, School of EECS, Peking University \and
		Zhejiang Lab \and 
		Department of Engineering Science, University of Oxford \and SenseTime Research  
	}
	\newcommand{\pp}{p}
	\newcommand{\FF}{\mathbf{F}}
	\newcommand{\XX}{\mathbf{X}}
	\newcommand{\YY}{\mathbf{Y}}
	
	\newcommand{\LXT}[1]{\textcolor{red}{{[\textbf{LXT}: #1]}}}
	
	\maketitle
	
\begin{abstract}
Existing semantic segmentation approaches either aim to improve the object's inner consistency by modeling the global context, or refine objects detail along their boundaries by multi-scale feature fusion.
In this paper, a new paradigm for semantic segmentation is proposed.
Our insight is that appealing performance of semantic segmentation requires \textit{explicitly} modeling the object \textit{body} and \textit{edge}, which correspond to the high and low frequency of the image.
To do so, we first warp the image feature by learning a flow field to make the object part more consistent.
The resulting body feature and the residual edge feature are further optimized under decoupled supervision by explicitly sampling different parts (body or edge) pixels.
We show that the proposed framework with various baselines or backbone networks leads to better object inner consistency and object boundaries.
Extensive experiments on four major road scene semantic segmentation benchmarks including \textit{Cityscapes}, \textit{CamVid}, \textit{KIITI} and \textit{BDD} show that our proposed approach establishes new state of the art while retaining high efficiency in inference. 
In particular, we achieve 83.7 mIoU \% on Cityscape with only fine-annotated data. 
Code and models are made available to foster any further research (\url{https://github.com/lxtGH/DecoupleSegNets}). 

\end{abstract}
	\section{Introduction}
Semantic segmentation is a fundamental task in computer vision that aims to assign an object class label to each pixel in an image.
It is a crucial step towards visual scene understanding, and has numerous applications such as autonomous driving~\cite{KITTI_dataset}, image generation~\cite{seg_generation_gan} and medical diagnosis.

Although the fully convolutional networks (FCNs)~\cite{fcn} have excelled in many major semantic segmentation benchmarks, they still suffer from the following limitations. 
First, the Receptive Field (RF) of FCNs grows slowly (only linearly) with increasing depth in the network, and such the limited RF is not able to fully model the longer-range relationships between pixels in an image~\cite{zhou2014object,luo2016understanding}. 
Thus the pixels are difficult to classify as the ambiguity and noise occurs inside the object body. 
Moreover, the downsampling operations in the FCNs lead to blurred predictions as the fine details disappear within the significantly lower resolution compared to the original image. 
As a result, the predicted segments tend to be blobby, and the boundary detail is far from satisfactory, which leads to a dramatic performance drop, especially on small objects.

\begin{figure}[!t]
	\centering
	\includegraphics[width=1.0\linewidth]{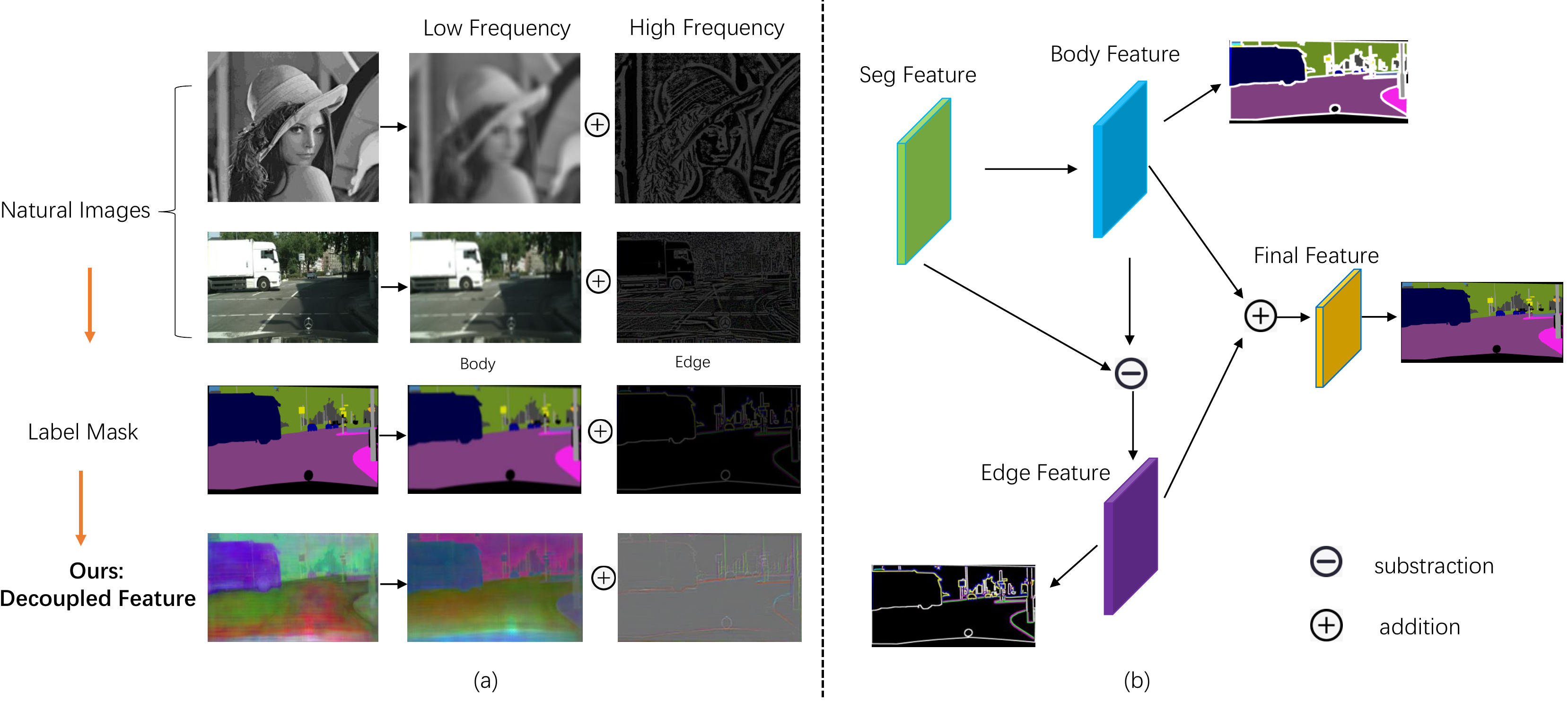}
	\vspace{-5mm}
	\caption{Illustration of our proposed module and supervision framework. (a). The motivation of our proposed framework. Our methods share the same splits with decoupling natural images into low frequency and high frequency. (b).Illustration of our framework. Our method consists of three steps: First, split segmentation feature into body feature and edge feature. Then both parts are supervised with specifically designed loss. Then merge both refined features for the final prediction.
	}
	\label{fig:teaser}
\end{figure}

To tackle the first problem, many approaches~\cite{dilation,pspnet,deeplabv3} have been proposed to enlarge the RF, such as dilated convolution, pyramid pooling module~\cite{deeplabv3,hou2020strip}, non-local operators~\cite{DAnet,ccnet,EMAnet,DMF_seg,li2019global}, graph convolution network (GCN)~\cite{beyond_grids,zhang2019dual} and dynamic graph~\cite{DGM_net}.
For the second problem, prior arts manage to embed the low-level features that contain boundary and edge information into high-level features~\cite{gated-scnn,Task_edge_detection,PGN_net,deeplabv3p} or directly refine the outputs~\cite{boundaries_network_fields}. 
However, the interaction between the object body and object edge is ignored. Can we solve both problems simultaneously?

On the other hand, it is natural for humans to distinguish objects by perceiving both object body and edge information. 
Inspired by this, we explore the relationships between body and edge in an explicit way to obtain the final semantic segmentation result. 
As shown in the first two rows of Fig.~\ref{fig:teaser}(a), a natural image can be decomposed into a low-spatial frequency component which describes the smoothly changing structure, and a high-spatial frequency component that represents the rapidly changing fine details. 
This is done by first applying mean or Gaussian filter for smoothing, and the remaining high-frequency parts can be obtained by subtraction. 
With the same philosophy, a segmentation mask can also be decoupled in this manner, where the finely detailed edge part can be obtained by subtraction from the body part. 
Inspired by this observation, we assume that a feature map for semantic segmentation can also be decoupled into two parts: body feature and edge feature (see Figure~\ref{fig:teaser}(b)). 
The former contains smooth representation inside the object with low frequency while the latter has sharper details with high frequency.

In this paper, we propose to solve semantic segmentation by explicitly modelling the body consistency and edge preservation in the feature level and then jointly optimizing them in a unified framework.

The entire process consists of three steps. 
First, we propose a novel flow-based method to generate body feature representation by warping each pixel towards object inner parts through a learned offset field to maintain the consistency of body part for each object.
Then, we obtain the edge feature by explicitly subtracting the body feature from the input feature.
The body feature is supervised by the mask where the edges are ignored during training, while the edge feature is supervised by an edge mask for learning edge prediction. 
Finally, we merge both optimized features into the final representation for segmentation. 
As the body generation part is done on a downsampled feature, the edge information is not accurate.
We follow the design of~\cite{video_propagation} to relax the object boundaries during body generation training, which makes both edge and body complementary to each other. 
Then both parts are merged into a single feature as a reconstructed representation, which is supervised by a commonly used cross-entropy loss. 
Moreover, the proposed framework is light-weighed and can be plugged into state-of-the-art FCNs~\cite{fcn,dilation,pspnet,deeplabv3p} based segmentation networks to improve their performance. 
Our methods achieve top performance on four driving scene semantic segmentation datasets including Cityscapes~\cite{Cityscapes}, CamVid~\cite{CamVid}, KITTI~\cite{KITTI_dataset} and BDD~\cite{yu2020bdd100k}. 
In particular, our method achieves 83.7 mIoU on Cityscapes datasets with only fine-annotated data. 

The contributions of this paper are as follows,
\begin{itemize}
\item  We propose a novel framework for the semantic segmentation task by decoupling the body and the edge with different supervisions. 
\item We propose a lightweight flow-based aggregation module by warping each pixel towards object inner parts through a learned offset field to maintain the consistency of body part for each object.
\item Our proposed module can be plugged into state-of-the-art segmentation methods to improve their performance with negligible cost. 
We carry out extensive experiments on four competitive scene parsing datasets and achieve top performance.
\end{itemize}
	\section{Related work}

\noindent \textbf{Semantic segmentation.} Recent approaches for semantic segmentation are predominantly based on FCNs~\cite{fcn}. Some earlier works~\cite{deeplabv1,deep_structured_seg,crf_as_rnn,seg_as_deep_parsing,crf_instance_seg,Sparse_filer_crf_} use structured prediction operators such as conditional random fields (CRFs) to refine the output boundaries. Instead of these costly DenseCRF, current state-of-the-art methods~\cite{pspnet,deeplabv3,DAnet,EMAnet} boost the segmentation performance by designing sophisticated head networks on dilated backbones~\cite{dilation} to capture contextual information. 
PSPNet~\cite{pspnet} proposes pyramid pooling module (PPM) to model multi-scale contexts, whilst DeepLab series~\cite{deeplabv2,deeplabv3,denseaspp} uses astrous spatial pyramid pooling (ASPP). In~\cite{DAnet,AdaptivePyramid_seg,EMAnet,annnet,ccnet,DMF_seg}, non-local operator~\cite{non_local} and self-attention mechanism~\cite{transformer} are adopted to harvest pixel-wise context from the whole image. Meanwhile, graph convolution networks~\cite{graph_cnn,beyond_grids,spg_net,DGM_net} are used to propagate information over the whole image by projecting features into an interaction space. Different from previous approaches, our method learns a flow field generated by the network itself to warp features towards object inner parts. 
DCN~\cite{deformable} uses predicted offset to aggregate information in kernel and SPN~\cite{spn_nips2017} proposes to propagate information through affinity pairs. 
Different from both work, our module aims to align pixels towards object inner according to the learned offset field to form body feature which is learned with specific loss supervision.
Ding \emph{et al}~\cite{BAFNet} models unidirectional acyclic graphs to propagate information within the object guided by the boundary. However, it is not efficient due to the usage of the RNN structure between pixels. Our module is light-weighted and can be plugged into the state-of-the-art methods~\cite{deeplabv3p,pspnet} to improve their performance with negligible extra cost, which also proves its efficiency and orthogonality. 

\noindent \textbf{Boundary processing.} Several prior works obtain better boundary localization by structure modeling, such as boundary neural fields~\cite{boundaries_network_fields}, affinity field~\cite{aaf}, random walk~\cite{cnn_random_wark}. The work~\cite{Task_edge_detection,PGN_net} uses edge information to refine network output by predicting edge maps from intermediate CNN layers. However, these approaches have some drawbacks, such as the potential error propagation from wrong edge estimation since both tasks are not orthogonal. Also overfitting edges brings noise and leads to inferior final segmentation results. Zhu et al.~\cite{video_propagation} proposes boundary relation loss to utilize coarse predicted segmentation labels for data augmentation. Inspired by the idea of label relaxation~\cite{video_propagation}, we supervise the edge and the body parts respectively. The relaxation body avoids the noise from the edge supervision with the relaxation loss. Experimental results demonstrate both higher model accuracy.

\noindent \textbf{Multi task learning.} Serveral works have proved the effectiveness of combining networks for complementary tasks learning~\cite{cross_stitch_net,ubernet}. The works of previous unified architectures that learn a shared representation using multi-task losses. There are some works~\cite{gated-scnn,PAD-net} using learned segmentation and boundary detection network simultaneously and the learned boundaries as an intermediate representation to aid segmentation. GSCNN~\cite{gated-scnn} designs a two-stream network by merging shape information into feature maps explicitly and introduces a dual-task loss to refine both semantic masks and boundary prediction. Different from these works, our goal is to improve the final segmentation results by explicitly optimizing two decoupled feature maps, and we design a specific framework by decoupling semantic body and boundaries into two orthogonal parts with corresponding loss functions and merge them back into final representation for segmentation task.
	\section{Method}
In this section, we will first introduce the entire pipeline of our proposed framework in Sec.~\ref{sec:framework}. Then we will describe the detailed description of each component in the Sec. 3.2-3.4. Finally, we present the network architectures equipped with our proposed modules and give some discussion on design in Sec.~\ref{sec:net_arch}.

\subsection{Decoupled segmentation framework}
\label{sec:framework}
Given a feature map $F \in \RR^{H \times W \times C}$, where $C$ represents the channel dimension and $H \times W$ means spatial resolution, our module outputs the refined feature map $\hat{F}$ with the same size. As stated in the introduction part, $F$ can be decoupled into two terms $F_{body}$ and $F_{edge}$. In this paper, we assume they meet the additive rule, which means $F$: $ F = F_{body} +F_{edge} $. Our goal is to design components with specific supervision to handle each parts, respectively. We achieve this by first performing body generation and then obtaining the edge part by explicit
subtraction where $F_{body} = \alpha(F)$ and $F_{edge} = F-F_{body}$. Then the refined feature $\hat{F}$ can be shown in $\hat{F} =  \phi(F) + \varphi(F_{edge})=  F_{body} + \varphi(F-F_{body})$. $\phi$ is the body generation module, which is designed to aggregate context information inside the object and form a clear body for each object. $\varphi$ represents the edge preservation module.
We will specify the details of $\phi$ and $\varphi$ in the following sections.

\begin{figure}[!t]
	\centering
	\includegraphics[width=0.85\textwidth]{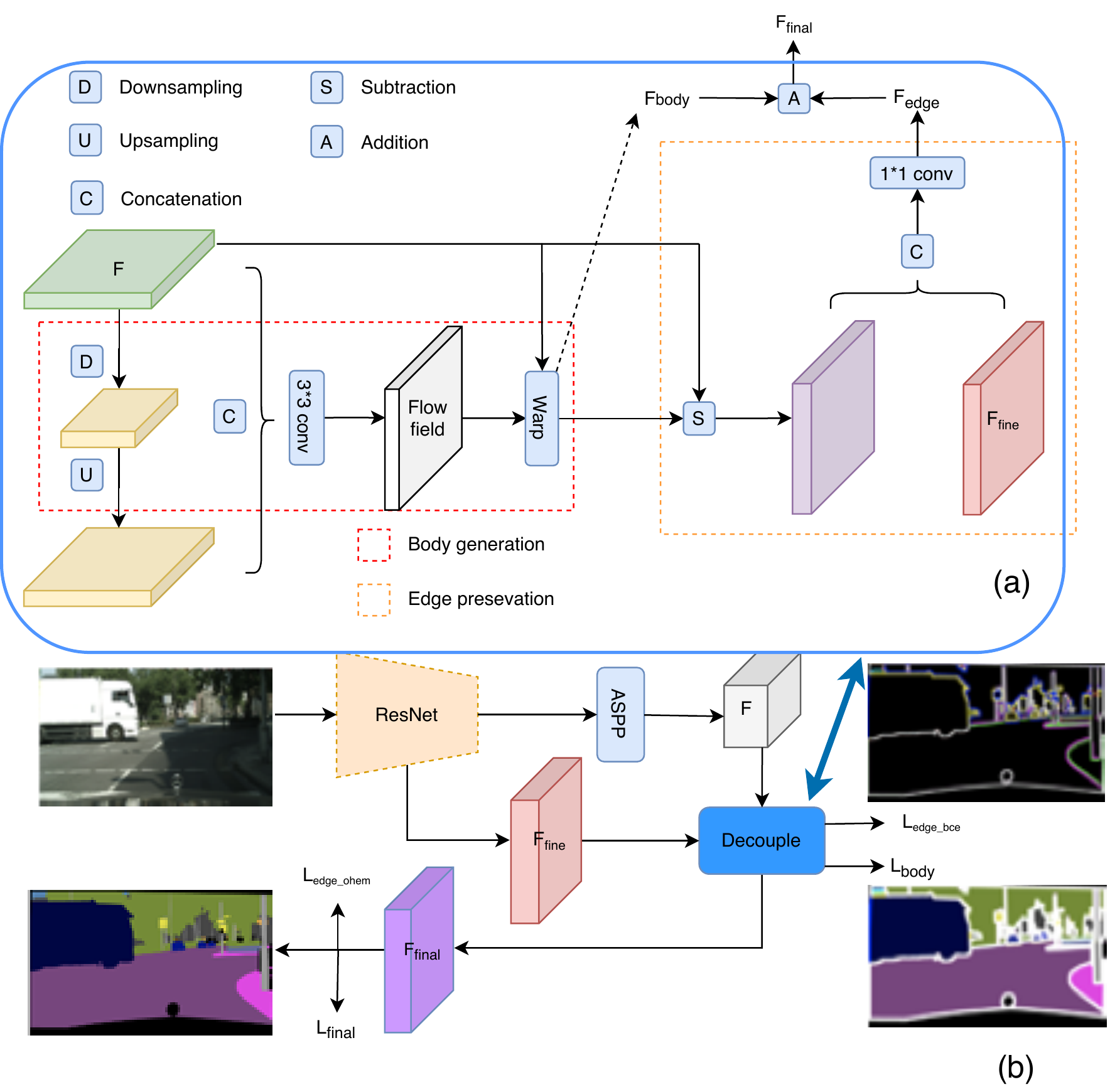}
	
	\caption{Illustration of our proposed module and supervision framework. (a) shows the proposed decoupled module with Body Generation and Edge Preservation. 
		(b) gives the examples of deploying our methods into Deeplabv3+~\cite{deeplabv3p}.}
	\label{fig:framework}
\end{figure}

\subsection{Body generation module}\label{sec:bg}
\label{sec:BG}
The body generation module is responsible for generating more consistent feature representations for pixels inside the same object. 
We observe that pixels inside an object are similar to each other, while those lying along the boundary show discrepancy.
We propose to explicitly learn body and edge feature representation.
To achieve so, we learn a flow field $\delta \in \RR^{H\times W \times 2}$, and use it to warp the original feature map to obtain the explicit body feature representation.
This module contains two parts: flow field generation and feature warping. 

\noindent
\textbf{Flow field generation.} 
To generate flows that mainly point towards object centers, it is a reasonable way to highlight the features of object center parts as explicit guidance.
Generally, low-resolution feature maps~(or coarse representation) often contain low-frequency terms.
Low spatial frequency parts capture the summation of images, and a lower resolution feature map represents the most salient part where we view it as pseudo-center location or the set of seed points. 
As shown in Fig.~\ref{fig:framework}(a), we adopt encoder-decoder design where the encoder downsamples feature into low-resolution representation with lower spatial frequency parts. We apply strided-convolutions to compress $F$ into the low-frequency map $F_{low}$. To be noticed, low resolution feature representation may also contain the high frequency information in some cases. However, we assume such condensed representation summarizes most salient parts on objects and leads to the coarse representation where the fine details or high frequency parts are ignored.
In particular, we adopt three successive $3 \times 3$ depthwise convolution to achieve that.  For flow field generation, we share the same pipeline as FlowNet-S~\cite{FlowNet}.
In detail, we first upsample $F_{low}$ to the same size as $F$ via bilinear interpolation, then concatenate them together and apply a $3 \times 3$ convolution layer to predict the flow map $\delta \in \RR^{H\times W \times 2}$.
Since our model is based on the dilated backbone network~\cite{dilation}, $3 \times 3$ kernel is large enough for covering the long distance between pixels in most cases.
More empirical improvements analysis on this implementation can be found in Sec.~\ref{sec:exp-city}. 

\noindent
\textbf{Feature warping.} 
Each position $\pp_{l}$ on standard spatial grid $\Omega_{l}$ is mapped to a new point $\hat{\pp}$ via $\pp_{l}+\delta_{l}(\pp_{l})$, we then use the differentiable bilinear sampling mechanism to approximate each point $\pp_{x}$ in $F_{body}$. The sampling mechanism, proposed in the spatial transformer networks~\cite{STN,DFF}, linearly interpolates the values of the four nearest neighbor pixel of $\pp_{l}$. The process is shown in Equation~\ref{equ:warpping}. 

\begin{equation}
F_{body}(p_{x}) = \sum_{\pp \in \mathcal{N}(\pp_{l})} w_\pp F(\pp)
\label{equ:warpping}
\end{equation}

\noindent
where $w_\pp$, calculated from flow map $\delta$, represents bilinear kernel weights on warped spatial gird. $\mathcal{N}$ represents the involved neighboring pixels. 

\subsection{Edge preservation module} 
\label{sec:EP}
The edge preservation module is designed to deal with high-frequency terms. It also contains two steps: 1) subtracting the body from the original feature $F$ and 2) adding fine-detailed lower-level features as a supplement. First, we subtract the body feature from the original input $F$. Inspired by recent works on decoder design~\cite{deeplabv3p}, we add extra low-level feature input as the supplement of missing fine details information to enhance the high-frequency terms in $F_{
	edge}$. Finally, we concatenate both and adopt a $1 \times 1$ convolution layer for fusion. This module can be formulated as Equation~\ref{equ:edge_preservation}, where $\gamma$ is a convolution layer and $||$ denotes the concatenation operation.

\begin{equation}
F_{edge} = \gamma ((F - F_{body}) || F_{fine}) 
\label{equ:edge_preservation}
\end{equation}

\subsection{Decoupled body and edge supervision}
\label{sec:DSL}

Instead of supervising the final segmentation map only, we jointly supervise all three parts, including $F_{body}$, $F_{edge}$ and $\hat{F}$, since each part has a specific purpose in our design. In particular, we append auxiliary supervised losses for $F_{body}$ and $F_{edge}$,receptively.
For the edge preservation module, we predict a boundary map $b$ during training, which is a binary representation of all the outlines of objects and stuff classes in the images. The total loss $L$ is computed as: 

\begin{equation}
L = \lambda_{1}L_{body}(s_{body},\hat{s}) + \lambda_{2}L_{edge}(b,s_{final},\hat{b},\hat{s}) + \lambda_{3} L_{final}(s_{final}, \hat{s})
\end{equation}

\noindent
where $\hat{s}$ represents the ground-truth~(GT) semantic labels and $\hat{b}$ is the GT binary masks which is generated by $\hat{s}$. $s_{body}$ and $s_{final}$ denote segmentation map prediction from $F_{body}$ and $F_{final}$ respectively. $\lambda_{1}$, $\lambda_{2}$, $\lambda_{3}$ are three hyper-parameters that control the weighting among the three losses and we set them $1$ as default. Note that $L_{final}$ is a common cross entropy loss for segmentation task and we detail the first two items as follows.

To make the optimization easier, for the $F_{body}$ part training, we relax the object boundaries during the training. We use the boundaries relaxation loss~\cite{video_propagation}, which only samples part of pixels within the objects for training.

For the edge part, we propose an integrated loss based on the boundaries edge prior which is got from edge prediction part. For semantic segmentation, most of the hardest pixels to classify lie on the boundary between object classes. Moreover, it is not easy to classify the center pixel of a receptive field when potentially half or more of the input context could be a new class. To tackle this problem, we propose to use such edge prior to handling the boundary pixels particularly and perform online hard example mining at a given edge threshold $t_b$ during the training. The total loss contains two terms: $L_{bce}$ is the binary cross-entropy loss for the boundary pixel classification, while $L_{ce}$ represents cross-entropy loss on edges parts in the scene. The formulations are shown in Equation~\ref{edge_loss} and Equation~\ref{eq:edge_ce_loss}.

\begin{equation}
\label{edge_loss}
L_{edge}(b,s,\hat{b},\hat{s}) = \lambda_{4} L_{bce}(b,\hat{b}) + \lambda_{5}L_{ce}(s,\hat{s},b)
\end{equation}

\begin{equation} 
\label{eq:edge_ce_loss}
L_{ce}(s,\hat{s},b) =  - \frac{1}{K} \sum_{i=1}^N w_i \cdot \mathbbm{1}[s_{i,\hat{s_i}} < t_K \cap \sigma(b_i) > t_b] \cdot \log{s_{i,\hat{s_i}}}
\end{equation}
For $L_{edge}$, we set $\lambda_{4}=25$ and $\lambda_{5}=1$ to balance the amount of pixels on edges.
For $L_{ce}$, we combine weighted bootstrapped cross-entropy with edge prior from $b$. 
We set $K = 0.10 \cdot N$, where $N$ is the total number of pixels in the image and $\hat{s}_i$ is the target class label for pixel $i$, $s_{i,j}$ is the predicted posterior probability for pixel $i$ and class $j$, and $\mathbbm{1}[x] = 1$ if $x$ is true and 0 otherwise. $\sigma$ represents the Sigmoid function to indicate whether $s$ is on the boundary. The threshold $t_K$ is set in a way that only the pixels with K highest losses are selected while the threshold $t_b$ is to mask non-boundary pixels. 
Both loss $L_{body}$ and $L_{edge}$ work complementary with each other by sampling pixels separately from different regions in the image. Such design benefits the final performance shown in the experimental parts.


\subsection{Network architecture}
\label{sec:net_arch} 
Fig.~\ref{fig:framework} illustrates the whole network architecture, which is based on state-of-the-art model Deeplabv3+~\cite{deeplabv3p}. Here we utilize dilated ResNet as backbone~\cite{resnet,dilation} only for illustration purpose. In particular, our module is inserted after the ASPP module~\cite{deeplabv3p}.
The decoupled supervisions are appended at the end of decouple module respectively. Moreover, our module is lightweight and can be deployed upon any FCN architectures such as PSPNet~\cite{pspnet} to refine the feature representation. When deployed on the native FCN, it is appended after the final output layer of the backbone. When deployed on the PSPNet, it is appended after the PPM module~\cite{pspnet}. $F_{fine}$ shares the same design with Deeplabv3+ for both architectures. 
	\section{Experiment}

\noindent
\textbf{Experiment settings and evaluation metrics:}
We first carry out experiments on the Cityscapes dataset, which comprises a large, diverse set of high resolution ($2048 \times 1024$) images recorded in street scenes. It consists of $5,000$ images with high-quality pixel-wise annotations for 19 classes, which is further divided into $2975$, $500$, and $1525$ images for training, validation and testing. To be noted, $20,000$ coarsely labeled images provided by this dataset are \textbf{not} used. Furthermore, we also evaluate our methods on CamVid~\cite{CamVid}, KITTI~\cite{KITTI_dataset} and BDD~\cite{yu2020bdd100k} datasets. For all datasets, we use the standard mean Intersection over Union (mIoU) metric to report segmentation accuracy. For Cityscapes, we also report F-score proposed in ~\cite{VOS_benchmark} by calculating along the boundary of the predicted mask given a small slack in the distance to show the high-quality segmentation boundaries of predicted mask.

\noindent
\textbf{Implementation details:} We use PyTorch~\cite{pytorch} framework to carry out the following experiments. All networks are trained with the same setting, where stochastic gradient `xsdescent (SGD) with a batch size of 8 is used as the optimizer, with the momentum of $0.9$, the weight decay of $5e-4$ and the initial learning rate of $0.01$. As a common practice, the `poly' learning rate policy is adopted to decay the initial learning rate by multiplying $(1 - \frac{\text{iter}}{\text{total}\_\text{iter}})^{0.9}$ during training. Data augmentation contains random horizontal flip, random resizing with scales range of $[0.75, ~2.0]$, and random cropping with size 832. Specifically, we use ResNet-50, ResNet-101~\cite{resnet} and Wider-ResNet~\cite{wide_resnet} as the backbones. Additionally, we re-implement the state-of-the-arts~\cite{pspnet,deeplabv3p} for fairness. We run the training for $180$ epochs for ablation purposes, and we run $300$ epochs for the submission to the test server. We first train the base networks without our module as initial weights and then train with our framework with the same epoch for each experiment. All the models are evaluated with the sliding-window manner for a fair comparison. 

\subsection{Ablation studies}\label{sec:exp-city}

\noindent
\textbf{Improvements over baseline model.} 
We first apply our method on naive dilated FCN models~\cite{dilation}, where we also include uniform sampling trick~\cite{video_propagation} to balance classes during training as our strong baselines in Table~\ref{tab:city_ablation}(a). Our naive FCNs achieve $76.6$ and $77.8$ in mIoU for ResNet-50 and ResNet-101, respectively. After applying our method, we achieve significant improvements over each backbone by $3.5\%$ and $3.0\%$, respectively. Note that our ResNet-50 based model is $2.2\%$ higher than ResNet-101 baseline, which indicates the performance gain is not from the more convolution layers in Body Generation.

\begin{table}[!t]
	\centering
	\begin{minipage}{\textwidth}
	    \centering
	    \begin{minipage}{\dimexpr.40 \linewidth}
			\centering
			\resizebox{0.85\textwidth}{!}{%
				\begin{tabular}{ l|c|c|c}
					\hline
					Method & Backbone &mIoU~(\%) & $\Delta$(\%)\\
					\hline
					FCN naive & ResNet-50 &75.4  & - \\
					\hline
					+ US~\cite{video_propagation}(Baseline)& ResNet-50 & 76.6  & - \\
					+ ours & ResNet-50 & \textbf{80.1} & \textbf{3.5}$\uparrow$\\
					\hline
					+ US~\cite{video_propagation}(Baseline)& ResNet-101 & 77.8  & - \\
					+ ours & ResNet-101 & \textbf{80.8} & \textbf{3.0}$\uparrow$\\
					\hline
				\end{tabular}
			}\par
			{ (a) Ablation study on strong FCN baselines.}
		\end{minipage}
		\medskip
	    \begin{minipage}{\dimexpr.45 \linewidth}		
			\centering
			\resizebox{0.85\textwidth}{!}{%
				\begin{tabular}{c c c c l l}
					\hline
					Method & $L_{body}$ & $L_{bce}$ & $L_{edge-ohem}$ & mIoU~(\%) & $\Delta$(\%) \\  
					\hline
					FCN  &  &  &  & 76.6 & -  \\ 
					+(BG \& EP)  & - & - & - & 78.3 & 1.7 $\uparrow$  \\ 
					\hline
					& \checkmark & - & - & 78.8 & 0.5$\uparrow$ \\
					& - & \checkmark & - & 78.3 & - \\
					& - &\checkmark &\checkmark & 78.7 & 0.4$\uparrow$\\
					&\checkmark &\checkmark &\checkmark & \textbf{80.1} & \textbf{1.8}$\uparrow$\\
					\hline
					w/o $F_{fine}$ &\checkmark &\checkmark &\checkmark & 79.3 & 0.8 $\downarrow$ \\
					w/o $ohem$ &\checkmark &\checkmark & $\times$ & 79.0 & 1.1 $\downarrow$ \\
					\hline
				\end{tabular}
			}\par
			{(b) Ablation study on Decoupled Supervision.}
		\end{minipage}
		
	\end{minipage}
	
	\begin{minipage}{\textwidth}
		\centering
		\begin{minipage}{\dimexpr.45 \linewidth}
		    \centering
			\resizebox{0.75\textwidth}{!}{%
				\begin{tabular}{ l|c|c}
					\hline
					Method & mIoU~(\%) & $\Delta $(\%)\\
					\hline
					FCN + BG \& EP (Baseline) & 78.3  & - \\
					\hline
					w/o BG warp & 76.9 & 1.4 $\downarrow$ \\
					w/o BG encoder-decoder & 77.3 & 1.0 $\downarrow$  \\ 
					w/o EP & 77.9 & 0.4 $\downarrow$ \\ 	
					\hline
				\end{tabular}
			}\par
			{(c) Ablation Study on effect of each component. }
		\end{minipage}
		\begin{minipage}{\dimexpr.45 \linewidth}
		    \centering
			\resizebox{0.65\textwidth}{!}{%
				\begin{tabular}{ l|c|c}
					\hline
					Method & mIoU~(\%) & $\Delta $(\%)\\
					\hline
					FCN (Baseline) & 76.6  & - \\
					\hline
					w SPN~\cite{spn_nips2017} & 77.9 & 1.3$\uparrow$ \\
					w DCN~\cite{deformable} & 78.2 & 1.6 $\uparrow$ \\ 			
					+GSCNN~\cite{gated-scnn} & 77.8 & 1.2$\uparrow$ \\
					ours & \textbf{80.1} & \textbf{3.5}$\uparrow$ \\
					\hline
				\end{tabular}
			}\par
			{(d) Comparison to related methods. }
		\end{minipage}
	\end{minipage}
	\caption{Experimental results on the Cityscapes validation set with dilated FCN as baselines.}
	
	\label{tab:city_ablation}
\end{table}

\noindent
\textbf{Ablation studies on decoupled supervisions.}
Then we explore the effectiveness of decoupled supervision in Table~\ref{tab:city_ablation}(b). Directly adding both the body generation and edge preservation module results in a 1.7\% improvement, which shows its aligned effect. After appending $L_{body}$, we get an obvious improvement of 0.5\%, and it can avoid uncertain noises on boundaries. $L_{bce}$ has no effect on the final performance since there is no direct supervision to segmentation prediction. Adding $L_{ce}$ will bring about 0.4\% improvement, which indicates that our integrated loss can better mine the boundaries based shape prior. Finally, after combining all three losses, we get a higher improvement by 1.8\%, which demonstrates the orthogonality of our separated supervision design. We also remove the $F_{fine}$ module in Equation~\ref{equ:edge_preservation}, which results in about 0.8\% drop in the final performance. This indicates the effectiveness of the edge cue from low-level features. Meanwhile, we also remove the hard pixel mining on $L_{edge-ohem}$, which results in about a 1.1\% drop. That shows the effectiveness of our proposed integrated loss on boundaries.

\noindent
\textbf{Ablation study on the effect of each component.} 
Here we carry out more detailed explorations on our component design with no the decoupled supervision setting shown in Table~\ref{tab:city_ablation}. Removing warping in BG achieves 76.9 in mIoU, which is a big decrease while removing the encoder-decoder part of BG results in 77.3\% due to the limited receptive field of dilated FCN. Removing EP leads to less performance drop as the main drop of FCN is on large objects shown in table~\ref{tab:city_improvement}.

\begin{table}[!t]
	\begin{minipage}{\textwidth}
		\begin{minipage}{\dimexpr.55\linewidth}		
			\centering
			\resizebox{0.70\textwidth}{!}{%
				\begin{tabular}{ l|c|c|c|c}
					\hline
					Method & Backbone & mIoU~(\%) & $\Delta $(\%)  & FLOPS \\
					\hline
					PSPNet~\cite{pspnet} & ResNet-50 & 79.6 & - & 132.1  \\
					+ours & ResNet-50 & 81.0 &   1.4$\uparrow$ & + 9.2~(6.8\%) \\
					Deeplabv3+~\cite{deeplabv3p} & ResNet-50 & 79.7 & - &  190.1\\
					+ours & ResNet-50 &  81.5 &  1.8$\uparrow$ & +9.0~(4.7\%) \\
					Deeplabv3+~\cite{EMAnet} & Wider-ResNet & 81.3 & - & 664.5  \\
					+ours & Wider-ResNet & 82.4 &  1.1$\uparrow$  & +7.5~(1.1\%) \\
					\hline
				\end{tabular}
			}\par
			{(a) Improvements upon different state-of-the-arts. To compute FLOPS, we adopt 512 $\times$ 512 images as the input.}	
		\end{minipage}
		\begin{minipage}{\dimexpr.45\linewidth}
			\centering
			\resizebox{0.75\textwidth}{!}{%
				\begin{tabular}{ l|c}
					\hline
					Method & mIoU(\%)  \\
					\hline
					Deeplabv3+~(ResNet-50) & 79.7 \\
					+ BG \& EP~(ResNet-50) &  81.5~(1.8$\uparrow$)  \\
					Deeplabv3+~(ResNet-101) & 80.7 \\
					+ BG \& EP~(ReseNet-101) & 82.6~(1.9$\uparrow$) \\
					+ BG \& EP~(ReseNet-101) +MS &  83.5 \\
					\hline
				\end{tabular}
			}\par
			{(b) Ablation study on improvement strategy on the validation set.}	
		\end{minipage}
	\end{minipage}
	
	\caption{Experiment results on Cityscapes validation set with more network architectures. Best viewed in color and zoom in.}
	\label{tab:city_improvement}
\end{table}

\begin{table}[!t]
	\centering
	\begin{minipage}{\dimexpr.40\linewidth}
		\centering
		\resizebox{0.8\textwidth}{!}{%
			\begin{tabular}{l|c|c}
				\hline
				Method & Backbone & mIoU~(\%)  \\
				\hline
				AAF~\cite{aaf}  & ResNet-101 &  79.1 \\ 
				PSANet~\cite{psanet} & ResNet-101 &  80.1 \\ 
				DFN~\cite{dfn} & ResNet-101 &  79.3 \\ 
				DenseASPP~\cite{denseaspp} & DenseNet-161 & 80.6 \\
				DAnet~\cite{DAnet} & ResNet-101 & 81.5 \\
				CCNet~\cite{ccnet} & ResNet-101 & 81.4 \\
				BAFNet~\cite{BAFNet} & ResNet-101 & 81.4 \\
				ACFNet~\cite{ACFNet} & ResNet-101 & 81.9 \\
				GFFnet~\cite{xiangtl_gff} & ResNet-101 & 82.3 \\
				\hline
				Ours & ResNet-101 & \textbf{82.8} \\
				\hline
			\end{tabular}
		}
		\par
		{(a) Results on Cityscapes test server trained with only fine-data.}
	\end{minipage}
	\begin{minipage}{\dimexpr.52 \linewidth}
		\centering
		
		\resizebox{0.8\textwidth}{!}{%
			\begin{tabular}{l|c|c|c}
				\hline
				Method & Coarse & Backbone & mIoU~(\%)  \\
				\hline
				PSP~\cite{pspnet} & \checkmark &ResNet-101 &  81.2 \\ 
				Deeplabv3+~\cite{deeplabv3p} & \checkmark& Xception & 82.1 \\
				DPC~\cite{DPC} & \checkmark & Xception & 82.6 \\
				Auto-Deeplab~\cite{auto-deeplab} & \checkmark & - & 82.1 \\
				Inplace-ABN~\cite{inplace_abn} & \checkmark & Wider-ResNet & 82.0 \\
				Video Propagation~\cite{video_propagation} & \checkmark & Wider-ResNet & 83.5 \\
				G-SCNN~\cite{gated-scnn} & $\times$ & Wider-ResNet &  82.8 \\
				\hline
				Ours  & $\times$ & Wider-ResNet & \textbf{83.7} \\
				\hline
			\end{tabular}
		}\par
		{(b) Results on the Cityscapes test server.}
		
	\end{minipage}
	\caption{Comparison with state-of-the-art on the Cityscapes test set. To be noted, our method \textbf{does not use coarse data.}}
	\label{tab:cityscapes_test_results}
\end{table}

\noindent
\textbf{Comparison with related methods.} To verify the effectiveness of the BG module, we replace our BG module with DCN and SPN operators. The former is used to aggregate features with learned offset field, while the latter propagate information through learned affinity pair. The first two rows in Table~\ref{tab:city_ablation}(f) demonstrates that our method works better than DCN and SPN, which proves the effectiveness of our BG module design. We also compare with GSCNN~\cite{gated-scnn} in the same setting. 

\noindent
\textbf{Improvements upon different base models.} To further verify the generality of our proposed framework, we test it upon several state-of-the-art models including PSPNet~\cite{pspnet} and Deelabv3+~\cite{deeplabv3p} with various backbone network in Table~\ref{tab:city_improvement}(a). It can be seen that our method improves those by around $0.9\%$-$1.5\%$ in mIoU. Note that our baselines are stronger than the original paper. Meanwhile, we also report the FLOPS during the inference stages in the last column of Table~\ref{tab:city_improvement}(a). Our module is extremely lightweight with only $1.1\%$-$6.8\%$ relative FLOPS increment.

\noindent
\textbf{Comparison to state-of-the-arts.} For fair comparison, we follow the common procedure of~\cite{DAnet,ccnet,gated-scnn} including stronger backbone~(ResNet-101~\cite{resnet}) and multi-scale inference~(MS) to improve the model performance. As shown in Table~\ref{tab:city_improvement}(b), our best model achieves 83.5 mIoU on the validation dataset after applying both techniques. Then we compare our method with state-of-the-arts on the Cityscapes test set in Table~\ref{tab:cityscapes_test_results} using the best model in Table~\ref{tab:city_ablation}. We first report results using ResNet-101 backbone in Table~\ref{tab:cityscapes_test_results}(a) and our method achieves \textbf{82.8} mIoU which improves by a large margin over all previous works. Moreover, we further apply our methods with a stronger backbone Wider-ResNet~\cite{wide_resnet} pretrained on the Mapillary~\cite{mapillary} dataset, which shares the same setting with GSCNN~\cite{gated-scnn}. Our method achieves \textbf{83.7} in mIoU and also leads to a significant margin over GSCNN~\cite{gated-scnn}. Table~\ref{tab:cityscapes_test_results}(b) shows the previous state-of-arts, which also uses much large coarse video data~\cite{video_propagation}), while our method achieves much better performance with utilizing only fine-annotated data.

\begin{table*}[!t]
	\centering 
	\small
	\addtolength{\tabcolsep}{0pt}
	\resizebox{\textwidth}{!}{
		\begin{tabular}{l|c|ccccccccccccccccccc}
			\hline
			Method  &  mIoU & road & swalk & build. & wall & fence & pole & tlight & sign & veg & terrain & sky & person & rider & car & truck & bus & train & motor & bike \\
			\hline
			FCN \cite{dilation} & 76.6 &98.0 & 84.5 &  92.5 & 50.7 & 62.7 &  \bd{67.7} & 73.8 & 81.2 & \bd{92.8} & 61.2  & \bd{94.7}  & \bd{83.8} & 64.2 & 95.0 & 56.4 & 81.6 & 60.5 & 68.2 & 79.4 \\ 
			Ours & \bd{80.1} & \bd{98.4} & \bd{86.4} &  \bd{92.9} & \bd{58.7} & \bd{64.8} &  67.0 & \bd{74.3} & \bd{82.2} & 92.7 & \bd{63.0}  & 94.5  & 83.6 & \bd{66.2} & \bd{95.2} & \bd{78.6} & \bd{91.0} & \bd{83.2} & \bd{69.3} & \bd{79.7} \\ 
			\hline
			PSPNet \cite{pspnet} & 79.6 & 98.0 & 84.5 & 92.9 & 54.9 & 61.9 &  66.5 & 72.2 & 80.9 & 92.6 & \bd{65.6}  & \bd{94.8}  & 83.1 & 63.5 & 95.4 & 83.9 & 90.6 & 84.01 & 67.6 & 78.5\\
			Ours & \bd{81.0} & \bd{98.2} & \bd{85.8} & \bd{93.4} & \bd{59.5} & \bd{67.0} &  \bd{68.7} & \bd{74.6} & \bd{81.6} & \bd{92.8} & 65.5  & 94.3  & \bd{83.6} & \bd{65.6} & \bd{95.6} & \bd{86.7} & \bd{92.6} & \bd{87.1} & \bd{68.6} & \bd{79.1} \\ 
			\hline
			Deeplabv3+ \cite{deeplabv3p} & 79.7 & 98.2 & 85.3 &  92.8 & 58.4 & 65.4 &  65.6 & 70.4 & 79.2 & 92.6 & 65.2  & 94.8  & 82.4 & 63.3 & 95.3 & 83.2 & 90.7 & 84.1 & 66.1 & 77.9 \\ 
			Ours & \bd{81.5} & \bd{98.3} & \bd{86.5} &  \bd{93.6} & \bd{60.7} & \bd{66.8} &  \bd{70.7} & \bd{73.9} & \bd{81.9} & \bd{93.1} & \bd{66.1}  & \bd{95.2}  & \bd{84.3} & \bd{67.5} & \bd{95.8} & \bd{86.1} & \bd{92.3} & \bd{85.5} & \bd{72.1} & \bd{80.1} \\ 
			\hline
		\end{tabular}
	}
	\caption{ 
		\small Per-category results on the Cityscapes validation set. Note that our method improves all strong baselines in most categories.
	} 
	\label{tab:cityscapes_results_detail_improvement}
\end{table*}


\subsection{Visual analysis}

\begin{figure}[!t]
	\centering
	\includegraphics[width=0.9\linewidth]{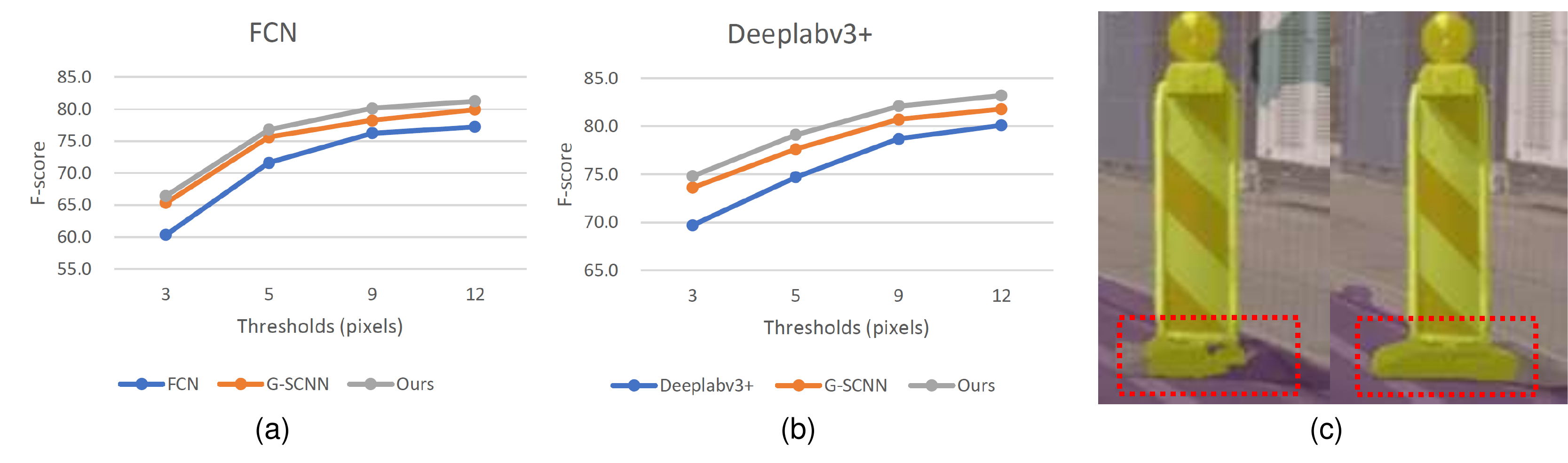}
	\caption{Improvement analysis on boundaries where F-score is adopted. (a) is the improvement on FCN. (b) is the improvement on Deeplabv3+. (c) is the improvement on mask boundary prediction. Best viewed in color and zoom in.
	}
	\label{fig:f_score}
\end{figure}

\begin{figure}[!t]
	\centering
	\includegraphics[width=0.8\linewidth]{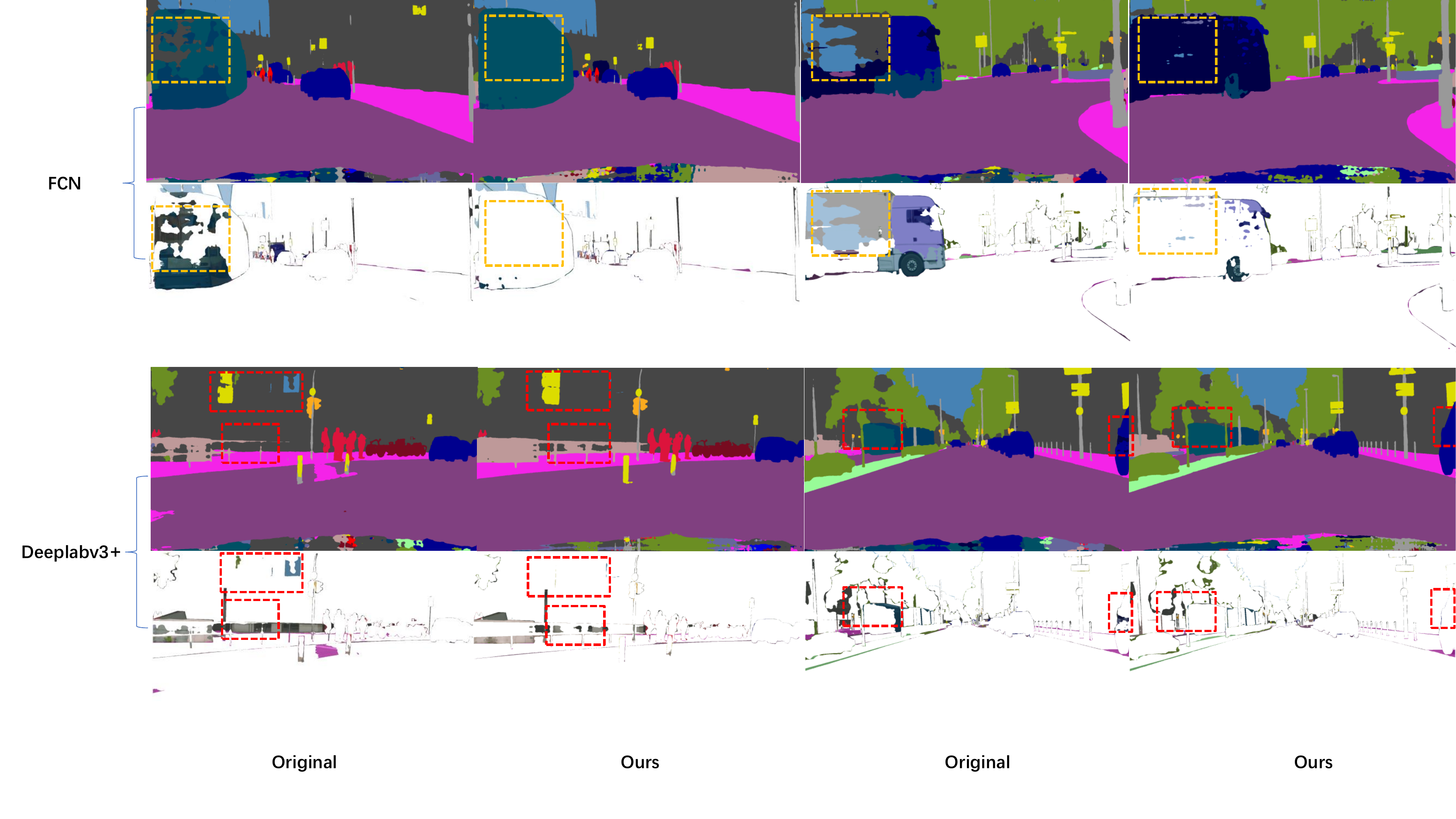}
	\caption{Comparisons upon on FCN and Deeplabv3+. The first and third rows are prediction masks while the second and last rows are error maps compared with the ground truth mask. The first two rows illustrate the results on FCN and ours, while the last two rows show Deeplabv3+'s and ours'. Our method solves the inner blur problem in large patterns for FCN shown in yellow boxes and fixes missing details and inconsistent results on small objects on Deeplabv3+ shown in red boxes. 
	}
	\label{fig:comparison_results}
\end{figure}

\noindent
\textbf{Improvement analysis.} Here we illustrate a detailed analysis of improvements. First we report mIoU of each category in Table~\ref{tab:cityscapes_results_detail_improvement}.
For the FCN model, our method improves a lot on large objects like bus and car in the scene. For Deeplabv3+ and PSPNet models, our method improves mainly on small objects such as traffic light and pole since most large patterns are handled by context aggregation modules like PPM and ASPP. To be more specific, we also evaluate the performance of predicted mask boundaries shown in Fig.~\ref{fig:f_score}, where we report the mean F-score of $19$ classes at 4-different thresholds. From that figure, we conclude that our methods improve the baseline object boundaries by a significant margin and our method is also slightly better than GSCNN~\cite{gated-scnn} on both different cases with four different thresholds. To be noted, we compared deeplabv3+ ResNet101 backbone in Fig.~\ref{fig:f_score}(b) with original paper while GSCNN results in Fig.~\ref{fig:f_score}(a) with the ResNet50 backbone implemented by us. Fig.~\ref{fig:f_score} (c) shows some visual examples of our model prediction with a more precise boundary mask. 
Fig.~\ref{fig:comparison_results} presents three visual examples over error maps. Our methods can better handle the inconsistency on large objects in FCN and boundaries of small objects in Deeplabv3+, which follows the same observation in Table~\ref{tab:cityscapes_results_detail_improvement}. 
More visual examples can be found in the supplementary file.

\begin{figure}[t]
	\centering
	\includegraphics[width=1.0\linewidth]{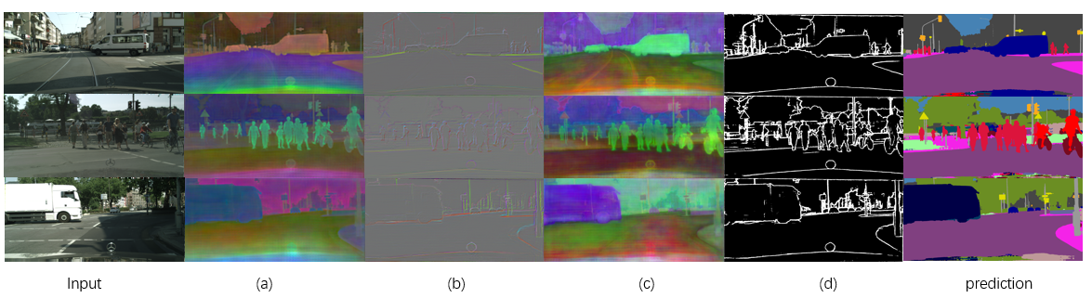}
	\caption{Visualization results based on Deeplabv3+ models. (a) is $F_{body}$. (b) is $F- F_{body}$. (c) is re-constructed feature $\hat{F}$. (d) is edge prior prediction $b$ with $t_{b}=0.8$. Best viewed in color and zoom in.
	}
	\label{fig:visualization}
\end{figure}

\noindent
\textbf{Visualization on decoupled feature representation and prediction.} 
We visualize the decoupled feature and prediction masks of our model over Deeplabv3+ in Fig.~\ref{fig:comparison_results}. Figures in (a)-(c) are drawn by doing Principal Component Analysis (PCA) from feature space into RGB space. As shown in Fig.~\ref{fig:visualization}, the features in (a) and (b) are complementary to each other, where each pixel in body part shares similar feature representations while pixels on edge varies. The merged feature in (c) have more precise and enhanced boundaries, while the objects in (a) are thinner than (c) due to boundary relaxation. The predicted edge prior in (d) has a more precise location of each object's boundaries. This gives better prior for mining hardest pixels along the boundaries parts. More visualization examples can be found in the supplementary file.

\begin{figure}[t]
	\centering
	\includegraphics[width=0.8\linewidth]{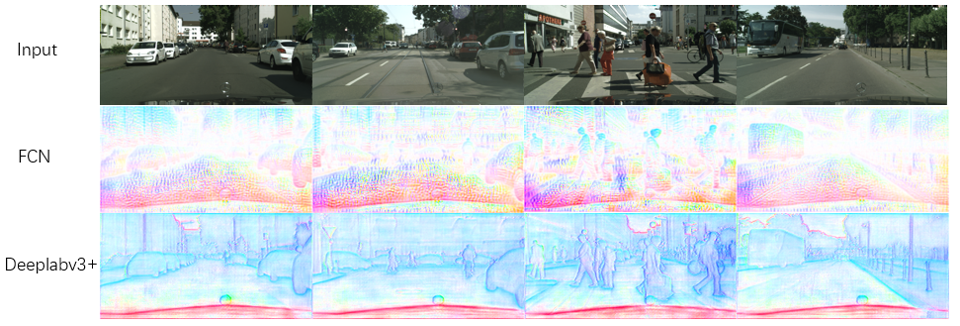}
	\caption{
		Flow maps visualization. The second row shows learned flow maps based on FCN, while the last row shows the learned maps based on Deeplabv3+.
	}
	\label{fig:flow_map}
\end{figure}

\noindent
\textbf{Visualization on flow field in BG.} We also visualize the learned flow field for FCNs and Deeplabv3+ in Fig.~\ref{fig:flow_map}. Both cases differ significantly. For the FCN model, we find that the learned flow field point towards the inner part in each object, which is consistent with our goal stated in Sec.~\ref{sec:bg}. While for the Deeplabv3+ model, the learned flow is sparse and mainly lies on the object boundaries because enough context has been considered in the ASPP module. This observation is consistent with the results in Table~\ref{tab:cityscapes_results_detail_improvement}: predictions over large objects are mainly improved in FCN~(truck, $22\%$), while those over small objects are mainly improved in Deeplabv3+~(pole, $5\%$).

\subsection{Results on other datasets}
To further prove the generality of our proposed framework, we also perform more experiments on the other three road sense datasets. Our model is the same as used in Cityscapes datasets, which is based on Deeplabv3+~\cite{deeplabv3p}. Standard settings of each benchmark are used, which are summarized in the supplementary file for the detailed reference.

\noindent
\textbf{CamVid:} CamVid~\cite{CamVid} is another road scene dataset. This dataset involves $367$ training images, $101$ validation images, and $233$ testing images with a resolution of $720\times960$. For a fair comparison, we compare both ImageNet pretrained and Cityscapes pretrained models. As shown in Table~\ref{tab:other_datasets}(a), our methods achieve significant gains over other state-of-the-arts in both cases.

\noindent
\textbf{BDD:} BDD~\cite{yu2020bdd100k} is a new road scene benchmark consisting $7,000$ images for training and $1,000$ images for validation. Compared with baseline model~\cite{yu2020bdd100k} which uses dilated backbone~($55.2\%$), our method leads to about $\mathbf{12\%}$ mIoU improvement with single scale inference with the same ResNet-101 backbone and achieves top performance with $66.9\%$ in mIoU.

\noindent
\textbf{KITTI:} KITTI benchmark~\cite{KITTI_dataset}  has the same data format as Cityscapes, but with a different resolution of $375\times1242$ and more metrics. The dataset consists of $200$ training and $200$ test images. Since it is a small dataset, we follow the settings from previous work~\cite{video_propagation,inplace_abn} by finetuning our best model from the Cityscapes dataset. Our methods rank first on three out of four metrics. It should be noted that method~\cite{video_propagation} uses both video and coarse data during Cityscapes pretraining process, while we only use the fine-annotated data.

\begin{table}[!t]
	\begin{minipage}{\textwidth}
	\centering
		\begin{minipage}{\dimexpr.45 \linewidth}
		    \centering
			\resizebox{0.8\textwidth}{!}{%
				\begin{tabular}{ l|c|c|c}
					\hline
					Method & Backbone & Pretrain & mIoU(\%)\\
					\hline
					DenseDecoder~\cite{densedecoder} & ResNext-101 & ImageNet & 70.9 \\
					BFP~\cite{BAFNet} & ResNet-101 & ImageNet & 74.1 \\
					Ours & ResNet-101 & ImageNet & \textbf{76.5}  \\
					\hline
					VideoGCRF~\cite{VideoGCRF} & ResNet-101 & Cityscapes & 75.2 \\
					Video Propagation~\cite{video_propagation} & Wider-ResNet & Cityscapes & 79.8 \\
  					Ours & ResNet-101 & Cityscapes & \textbf{81.5} \\
					Ours & Wider-ResNet & Cityscapes & \textbf{82.4} \\
					\hline
				\end{tabular}
			}\par
			{(a) Results on CamVid.}
		\end{minipage}	
		\medskip
		\begin{minipage}{\dimexpr.40 \linewidth}
		    \centering
			\resizebox{0.75\textwidth}{!}{%
			\begin{tabular}{ l|c|c}
					\hline
					Method & Backbone & mIoU(\%) \\
					\hline
					Dilated~\cite{yu2020bdd100k} & ResNet101 & 55.2 \\
					FasterSeg~\cite{fasterseg} & - & 55.3 \\ 
				Ours & ResNet101 & \textbf{66.9} \\
				\hline
				\end{tabular}	
			}\par
			
			{(b) Results on BDD.}
		\end{minipage}
    \end{minipage}
    
		\begin{minipage}{\dimexpr.90 \linewidth}
		    \centering
			\resizebox{0.65\textwidth}{!}{%
				\begin{tabular}{ l|c|c|c|c}
					\hline
					Method & IoU class(\%) & iIoU class(\%) & IoU category(\%) & iIoU category(\%) \\
					\hline
					AHiSS~\cite{AHiSS} & 61.2 & 26.9 & 81.5 & 53.4 \\
					LDN~\cite{LDN} & 63.5 & 28.3 & 85.3 & 59.1 \\ 
					MapillaryAI~\cite{inplace_abn} &69.6 & 43.2 & 86.5 & 68.9\\ 
					Video Propagation~\cite{video_propagation} & \textbf{72.8} & 48.7 & \textbf{88.9} & 75.2 \\ 
					Ours & \textbf{72.8} & \textbf{49.5} & 88.5 & \textbf{75.5} \\
					\hline
				\end{tabular}
			}\par
			{(c) Results on KITTI.}
		\end{minipage}

	\caption{Experiments results on other road scene benchmarks.}
	\label{tab:other_datasets}
\end{table}

	\section{Conclusions}
In this paper, we propose a novel framework to improve the semantic segmentation results by decoupling features into the body and the edge parts to handle inner object consistency and fine-grained boundaries jointly. We propose the body generation module by warping feature towards objects' inner parts then the edge can be obtained by subtraction. Furthermore, we design decoupled loss by sampling pixels from different parts to supervise both modules' training. Both modules are light-weighted and can be deployed into the FCN architecture for end-to-end training. We achieve state-of-the-art results on four road scene parsing datasets, including Cityscapes, CamVid, KITTI and BDD. The superior performance demonstrates the effectiveness of our proposed framework.

	\section{Acknowledgement}
	We gratefully acknowledge the support of SenseTime Research for providing the computing resources. Z. Lin is supported by NSF China (grant no.s 61625301 and 61731018) and Major Scientific Research Project of Zhejiang Lab (grant no.s 2019KB0AC01 and 2019KB0AB02).

\section{More Experimental Details}

We will give more detailed results on Cityscapes in this supplemental material.

\noindent
\textbf{More implementation details on the edge supervision:} As discussed in the paper, we use the edge supervision for two purposes: one for edge prediction and one for edge hardest pixel mining. For edge prediction, we use binary edge map generated from the category label map as the ground-truth, and add extra weights by counting the reciprocal of positive and negative pixels to balance the binary cross-entropy loss. For edge hardest pixel mining, we set $K=0.1$ based on the input resolution due to the various sizes in different datasets. Since our method requires a fine-grained edge from the ground truth mask, we only use the fine-annotated set.

\vspace{2ex}
\noindent
\textbf{More implementation details on related methods:} For SPN~\cite{spn_nips2017}, we use the authors' original Pytorch code~\cite{spn_nips2017} and append it after the FCN output to replace our proposed Body Generation module.
For DCN~\cite{deformable}, we use the implementation of mmdetction repo~\cite{mmdetection2018} and replace our Body Generation module with two DCN operators. Note that both supervisions and the edge preservation module are kept untouched. For G-SCNN~\cite{gated-scnn}, we port the author's open sourced code~\cite{gated-scnn} into our framework with extra shape stream and dual-task loss on the FCN.

\noindent
\textbf{Ablation study on component design in body generation~(BG).} 
Here we carry out more detailed explorations on our BG design. We adopt the same setting shown in the experiment part. Table~\ref{tab:city_ablation2}(a) shows that depth-wise conv works better than bilinear and is also slightly better than naive conv with less computation. Table~\ref{tab:city_ablation2}(b) shows that naive bilinear upsampling works better than deconvolution and nearest neighbor during upsampling. Table~\ref{tab:city_ablation2}(c) shows that two successive strided-conv with total stride 4 works the best while larger stride leads to degradation due to the loss of details. 

\vspace{2ex}
\noindent
\textbf{Detailed boundaries improvements analysis:} 
We analyze the improvements over boundaries using F-score~\cite{VOS_benchmark} on the Cityscapes dataset. The analysis is performed on Deeplabv3+ over each class with 4 different boundary thresholds.
For a fair comparison, we also include the G-SCNN~\cite{gated-scnn} results on the val set since both models are based on Deeplabv3+. The results are shown in Tab.~\ref{tab:cityscapes_results_detail_f_score}.

\vspace{2ex}
\noindent
\textbf{Detailed results on the Cityscapes test server:} 
We first give the comparison results with models trained with only fine-annotated data using ResNet-101 as the backbone in Tab.~\ref{tab:cityscapes_results_detail_fine}. 
Our method leads to a significant margin with previous state-of-the-art models and outperforms them in 18 of 19 categories. Then we compare the our model with Wider-ResNet in Tab.~\ref{tab:cityscapes_results_detail_large}. For both cases, we achieve state-of-the-art results.

\begin{table}[!t]
	\begin{minipage}{\textwidth}
	
	\vspace{2mm}	
	\begin{minipage}{\textwidth}
		\centering
		\begin{minipage}{\dimexpr.45 \linewidth}
		    \centering
			\resizebox{0.65\textwidth}{!}{%
				\begin{tabular}{ l|c|c}
					\hline
					Method & mIoU~(\%) & $\Delta $(\%)\\
					\hline
					FCN (Baseline) & 76.6  & \\
					\hline
					bilinear & 78.2 & 1.6$\uparrow$ \\
					naive conv & 79.7 & 3.1$\uparrow$ \\
					depth-wise-conv & \textbf{80.1} & \textbf{3.5}$\uparrow$ \\
					\hline
				\end{tabular}
			}\par
			{\footnotesize(a) Ablation study on downsampling operations in BG.}
			
		\end{minipage}	
		\begin{minipage}{\dimexpr.45 \linewidth}
		    \centering
			\resizebox{0.65\textwidth}{!}{%
				\begin{tabular}{ l|c|c}
					\hline
					Method & mIoU~(\%) & $\Delta $(\%)\\
					\hline
					FCN (Baseline) & 76.6 & - \\
					\hline
					de-conv & 79.0 & 2.4$\uparrow$ \\
					nearest neighbor & 79.5 & 2.9$\uparrow$ \\
					bilinear & \textbf{80.1} & \textbf{3.5}$\uparrow$ \\
					\hline
				\end{tabular}
			}\par
			{\footnotesize(b) Ablation study on upsampling operations in BG.}
		\end{minipage}
	\end{minipage}
	
	\end{minipage}
	
	\vspace{2mm}
	\begin{minipage}{\textwidth}
		\centering
		\begin{minipage}{\dimexpr.40 \linewidth}
		    \centering
			\resizebox{0.65\textwidth}{!}{%
				\begin{tabular}{ l|c|c}
					\hline
					Method & mIoU~(\%) & $\Delta $(\%)\\
					\hline
					FCN (Baseline) & 76.6  & - \\
					\hline
					(1, stride=2) & 79.2 & 2.6$\uparrow$  \\
					(2, stride=4) & \textbf{80.1} & \textbf{3.5}$\uparrow$\\ 			
					(3, stride=8) & 78.8 & 2.2$\uparrow$\\
					(4, stride=16) & 78.5 & 1.9$\uparrow$\\
					\hline
				\end{tabular}
			}\par
			{\footnotesize(c) Ablation study on number of strided-convs in BG.}
		\end{minipage}
	
	\end{minipage}

	\vspace{3mm}
	\caption{Experiment results on Cityscapes validation set with component design in body generation part.}
	\vspace{-5mm}
	\label{tab:city_ablation2}
\end{table}

\begin{table}[t!]
	\centering 
	\small
	\addtolength{\tabcolsep}{0pt}
	\resizebox{\textwidth}{!}{
		\begin{tabular}{l|c|c|ccccccccccccccccccc}
			\hline
			Method & Thrs & mIoU &  road & swalk & build. & wall & fence & pole & tlight & sign & veg & terrain & sky & person & rider & car & truck & bus & train & motor & bike \\
			\hline
			Deeplabv3+ & 3px & 69.7 & 83.7 & 65.1 & 69.7 & 52.2 & 46.2 &  72.2 & 62.8 & 67.7 & 71.8 & 52.2  & 80.9  & 61.5 & 66.4 & 78.8 & 78.2 & 83.9 & 91.7 & 77.9 & 60.9 \\ 
			G-SCNN & 3px & 73.6 & 85.0 & 68.8 &  74.1 & 53.3 & 47.0 &  79.6 & 74.3 & 76.2 & 75.3 & 53.1  & 83.5  & 69.8 & 73.1 & 83.4 & 75.8 & 88.0 & 93.9 & 75.1 & 68.5 \\ 
			Ours & 3px & \textbf{73.8} & 85.2 & 69.1 &  74.0 & 50.3 & 50.2 &  78.6 & 74.6 & 75.2 & 75.1 & 55.3  & 81.5  & 70.2 & 72.1 & 82.4 & 76.3 & 89.1 & 92.8 & 76.2 & 69.5 \\ 
			\hline
			Deeplabv3+ & 5px & 74.7 & 88.1 & 72.6 & 78.1 & 55.0 & 49.1 & 77.9 &
            69.0 & 74.7 & 81.0 & 55.8 & 86.4 & 69.0 &
            71.9 & 85.4 & 79.4 & 85.4 & 92.1 & 79.4 & 68.4 \\  
			G-SCNN & 5px & {77.6} &{88.7} & 75.3 & 80.9 & {55.9} & {49.9} &{83.6} & {78.6} & {80.4} & {83.4} & {56.6} & {88.4} & {75.4} &
            {77.8} & {88.3} & 77.0 & {88.9} & {94.2} & 76.9 & {75.1} \\
			Ours & 5px & \textbf{79.2} & 88.6 & 74.6 &  81.8 & 55.2 & 55.3 &  83.3 & 80.0 & 80.6 & 82.9 & 60.3  & 88.2  & 75.4 & 79.5 & 89.2 & 83.6 & 92.8 & 96.3 & 80.9 & 75.5 \\
			\hline
			Deeplabv3+ & 9px & 78.7 & 91.2 & 78.3 & 84.8 & 58.1 & 52.4 & 82.1 &
            73.7 & 79.5 & 87.9 & 59.4 & 89.5 & 74.7 &
            76.8 & 90.0 & {80.5} & 86.6 & 92.5 & {81.0} & 75.4 \\ 
			G-SCNN & 9px & {80.7} & {91.3} & {80.1} & {86.0} & {58.5} & {52.9} & {86.1} &
            {81.5} & {83.3} & {89.0} & {59.8} & {91.1} & {79.1} &
            {81.5} & {91.5} & 78.1 & {89.7} & {94.4} & 78.5 & {80.4} \\ 
			Ours & 9px & \textbf{82.3} & {91.5} & 79.7 & 87.4 & 57.7 & 58.3 & 86.1 & 83.1 & 83.8 & 88.9 & 63.7 & 90.8 & 79.3 & 83.5 & 92.5 & {84.6} & 93.5 & 96.6 & {82.4} & 82.4 \\
			\hline
			Deeplabv3+ & 12px & 80.1 & {92.3} & 80.4 & 87.2 & 59.6 & 53.7 & 83.8 &
            75.2 & 81.2 & 90.2 & 60.8 & 90.4 & 76.6 &
             78.7 & 91.6 & {81.0} & 87.1 & 92.6 & {81.8} & 78.0 \\ 
			G-SCNN & 12px & {81.8} & 92.2 & {81.7} & {87.9} & {59.6} & {54.3} & {87.1} &
            {82.3} & {84.4} & {90.9} & {61.1} & {91.9} & {80.4} &
            {82.8} & {92.6} & 78.5 & {90.0} & {94.6} & 79.1 & {82.2} \\ 
			Ours & 12px & \bf{83.5} & {92.4} & {81.5} & {89.4} & {58.8} & 59.5 & 87.1 & 83.9 & 84.9 & 91.0 & {65.0} & 91.6 & {80.6} & {84.9} & {93.5} & 85.1 & 93.7 & {96.7} & 82.9 & 83.1 \\
			\hline
		\end{tabular}
		}
	\vspace{2mm}
	\caption{ 
		\small Per-category F-score results on the Cityscapes val set for 4 different thresholds based on Deeplabv3+. Note that our methods output G-SCNN over \textbf{all four thresholds}. Best view on screen and zoom in.
	} 
	\label{tab:cityscapes_results_detail_f_score}
\end{table}

\begin{table}[t!]
\centering 
\small
\addtolength{\tabcolsep}{0pt}
    \resizebox{\textwidth}{!}{
	\begin{tabular}{l|c|ccccccccccccccccccc}
		\hline
		Method & mIoU &  road & swalk & build. & wall & fence & pole & tlight & sign & veg & terrain & sky & person & rider & car & truck & bus & train & motor & bike \\
		\hline
		PSPNet~\cite{pspnet} & 78.4 & 98.6 & 86.2 & 92.9 & 50.8 & 58.8 & 64.0 & 75.6 & 79.0 & 93.4 & 72.3 & 95.4 & 86.5 & 71.3 & 95.9 & 68.2 & 79.5 & 73.8 & 69.5 & 77.2 \\
		AAF~\cite{aaf} & 79.1 & 98.5 & 85.6 & 93.0 & 53.8 & 58.9 & 65.9 & 75.0 & 78.4 & 93.7 &
		72.4 & 95.6 & 86.4 & 70.5 & 95.9 & 73.9 & 82.7 & 76.9 & 68.7 & 76.4\\
		DenseASPP~\cite{denseaspp}  & 80.6 & \bf{98.7} & 87.1 & 93.4 & 60.7 & 62.7 & 65.6 & 74.6 & 78.5 & 93.6 & 72.5 & 95.4 & 86.2 & 71.9 & 96.0 & 78.0 & 90.3 & 80.7 & 69.7 & 76.8 \\
		DANet~\cite{DAnet} & 81.5 & 98.6 & 87.1 & 93.5 & 56.1 & \textbf{63.3} & 69.7 & 77.3 & 81.3 & 93.9 & 72.9 & 95.7 & 87.3 & 72.9 & 96.2 & 76.8 & 89.4 & 86.5 & 72.2 & 78.2 \\
		\hline
		Ours & \bf{82.8} & \bf{98.7} & \bf{87.2} & \bf{93.9} & \bf{62.1} & 62.9 & \bf{71.2} & \bf{78.5} & \bf{81.8} & \bf{94.0} & \bf{73.3} & \bf{96.0} & \bf{88.1} & \bf{74.4} & \bf{96.5} & \bf{79.4} & \bf{92.5} & \bf{89.8} & \bf{73.3} & \bf{78.7} \\
		\hline
	\end{tabular}
	}
\vspace{2mm}
\caption{\small{
		Per-category results on the Cityscapes test set. Note that all the models are trained with \textbf{only fine annotated data}. Our method outperforms existing approaches on \textbf{18} out of 19 categories, and achieves \textbf{82.8}\% in mIoU.}
}
\label{tab:cityscapes_results_detail_fine}
\end{table}

\begin{table}[t!]
	\centering 
	\small
	\addtolength{\tabcolsep}{0pt}
		\resizebox{\textwidth}{!}{
		\begin{tabular}{l|c|c|ccccccccccccccccccc}
			\hline
			Method & Coarse & mIoU &  road & swalk & build. & wall & fence & pole & tlight & sign & veg & terrain & sky & person & rider & car & truck & bus & train & motor & bike \\
			\hline
			PSP-Net \cite{pspnet} & \checkmark & 81.2 & 98.7 & 86.9 &  93.5 & 58.4 & 63.7 &  67.7 & 76.1 & 80.5 & 93.6 & 72.2  & 95.3  & 86.8 & 71.9 & 96.2 & 77.7 & 91.5 & 83.6 & 70.8 & 77.5 \\ 
			DeepLabV3 \cite{deeplabv3} & \checkmark & 81.3 & 98.6 & 86.2 & 93.5 & 55.2 & 63.2 & 70.0& 77.1 & 81.3 & 93.8    & 72.3&   95.9  &  87.6  & 73.4& 96.3 & 75.1 &  90.4  & 85.1 & 72.1 & 78.3 \\
			DeepLabV3+ \cite{deeplabv3p} & \checkmark & 81.9 & 98.7 & 87.0  & 93.9 & 59.5 & 63.7 & 71.4 &78.2 & 82.2& 94.0& 73.0 & 95.8&88.0&  73.3 & 96.4  &  78.0 & 90.9 & 83.9 & 73.8 & 78.9 \\
			AutoDeepLab-L \cite{auto-deeplab} & \checkmark & 82.1 & \bf{98.8} & 87.6  & 93.8  & 61.4  & 64.4  & 71.2 & 77.6 & 80.9 & 94.1 & 72.7 & 96.0 & 87.8 &  72.8  & 96.5  & 78.2 & 90.9 & 88.4 & 69.0 & 77.6 \\
			DPC \cite{DPC} & \checkmark & 82.7 & 98.7 & 87.1 & 93.8 & 57.7 & 63.5 & 71.0 & 78.0 & 82.1 &94.0 & 73.3 & 95.4 & 88.2 &  \bf{74.5}  & {96.5}  &  \bf{81.2} & {93.3} & {89.0} & {74.1} & 79.0 \\
			\hline       
			G-SCNN~\cite{gated-scnn} & $\times$ & 82.8 & 98.7 & 87.4 & 94.2 & 61.9 & \bf{64.6} & \bf{72.9} & \bf{79.6} & {82.5} & \bf{94.3} & \bf{74.3}& \bf{96.2} & 88.3 & 74.2 & 96.0 & 77.2& 90.1 & 87.7& 72.6& {79.4} \\
			Ours & $\times$ & \bf{83.7} & \bf{98.8} & \bf{87.8} & \bf{94.4} & \bf{66.1} & 64.7 & 72.3 & 78.8 & \bf{82.6} & 94.2 & 73.9 & 96.1 & \bf{88.6} & \bf{75.9} & \bf{96.6} & 80.2 &\bf{93.8} & \bf{91.6} & \bf{74.3} & \bf{79.5} \\
			\hline
		\end{tabular}
		}
	\vspace{2mm}
	\caption{ 
		\small Per-category results on the Cityscapes test set. Note that G-SCNN and our method are trained with \textbf{only fine annotated data}. We achieve the state-of-the-art results with \textbf{83.7} mIoU. Best view on screen and zoom in.
	} 
	\label{tab:cityscapes_results_detail_large}
\end{table}

\section{More Visualisation Results}
In this section, we give more visualization examples, as shown in the paper's Experiment parts.

\begin{figure}[!t]
	\centering
	\includegraphics[width=1.0\linewidth]{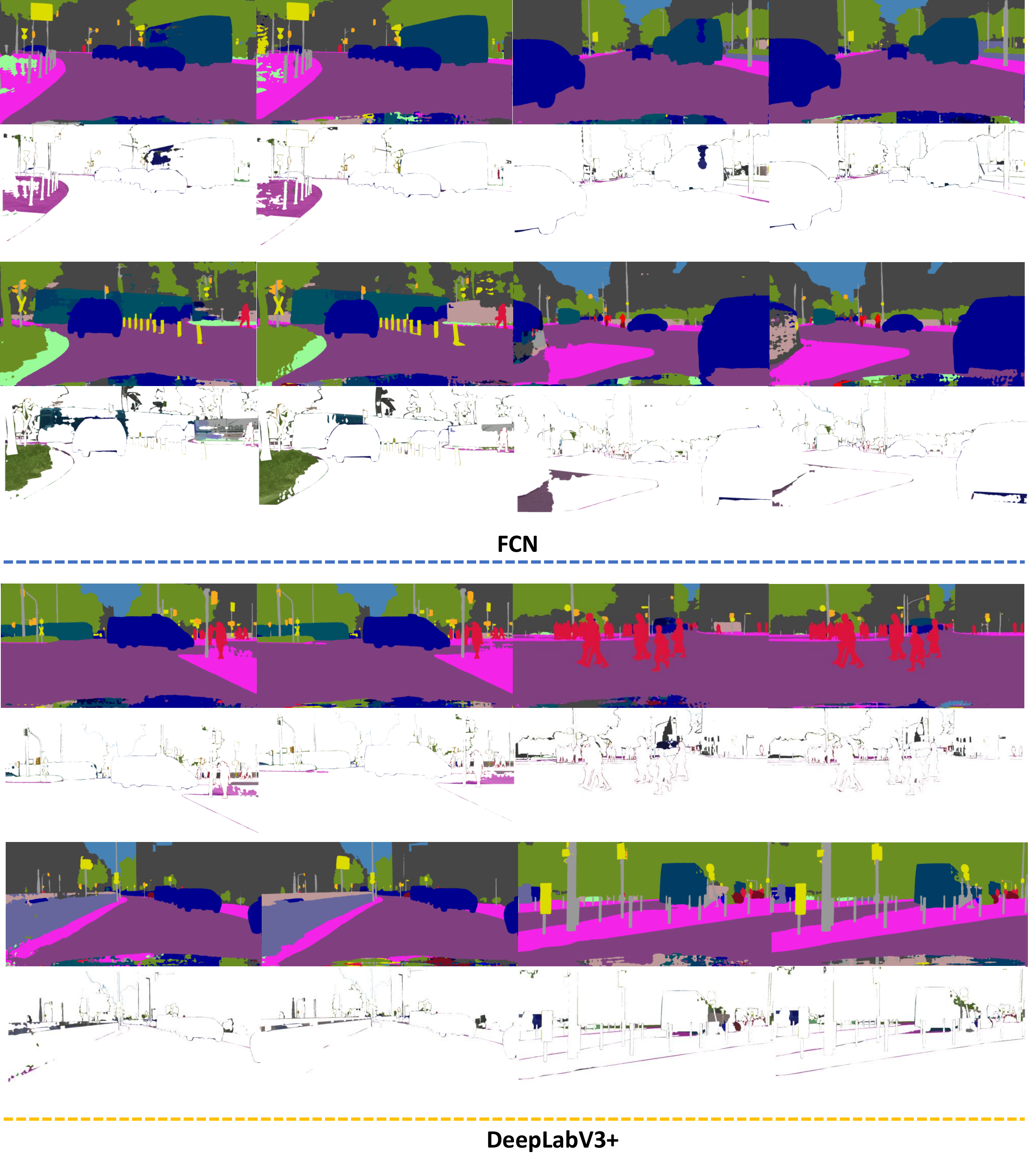}
	\vspace{-5mm}
	\caption{Improvements over FCN~(First four rows) and Deeplabv3+~(Last four rows). 
	Best view it in color and zoom in.}
	\label{fig:error_map_sub}
\end{figure}

\noindent
\textbf{More visualization improvement analysis:} In Fig.~\ref{fig:error_map_sub}, we include more visual comparisons on FCN and Deeplabv3+ with our methods. The right figures are our method's outputs. Our method solves the inner blur
problem in large patterns for FCN and fixes missing details and inconsistent results on small objects on Deeplabv3+.

\begin{figure}[!t]
	\centering
	\includegraphics[width=1.0\linewidth]{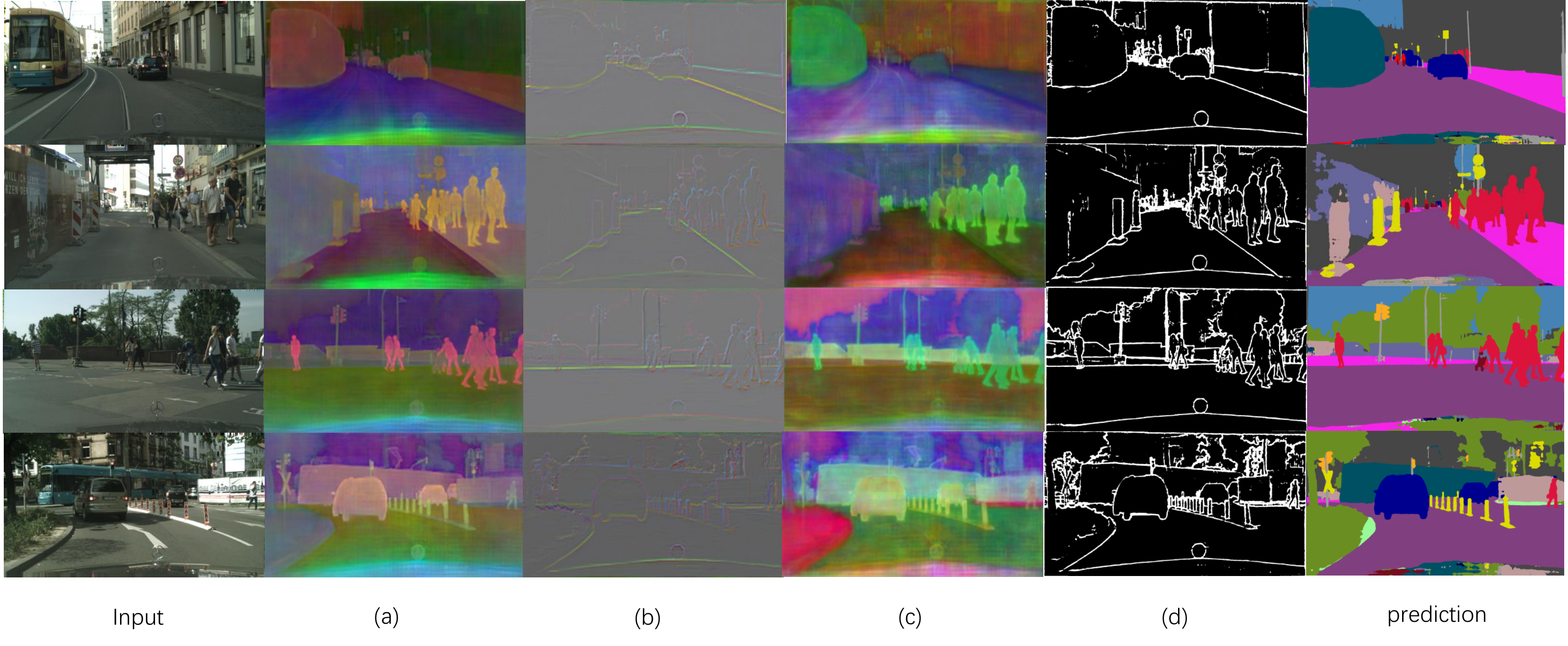}
	\vspace{-5mm}
	\caption{More examples on Decoupled Feature Representation.(a) is $F_{body}$. (b) is $F- F_{body}$. (c) is re-constructed feature $\hat{F}$. (d) is edge prior prediction $b$ with $t_{b}=0.8$. Best view it in color and zoom in.}
	\label{fig:decouple_fea_sub}
\end{figure}

\vspace{2ex}
\noindent
\textbf{More visualization on decoupled feature representations and predictions:}
We give more visualization examples on decouple feature representation in Fig.~\ref{fig:decouple_fea_sub}.

\begin{figure}[t]
	\centering
	\includegraphics[width=1.0\linewidth]{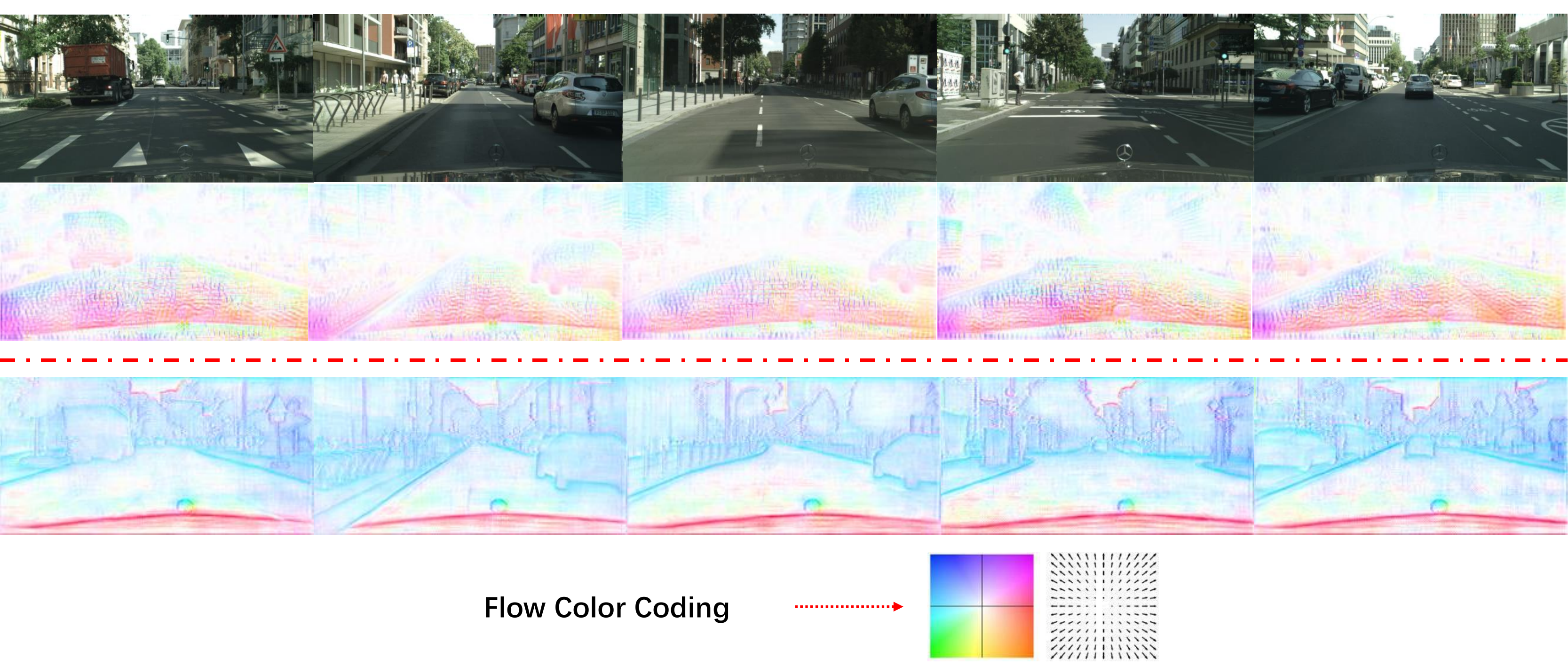}
	\vspace{-5mm}
	\caption{
		Flow field visualizations. The first row shows the input images.
		The second row shows the generated flow fields based on FCN, while the third row shows the generated fields based on Deeplabv3+. We show flow directions and the color map in the last row.
	}
	\label{fig:flow_map_sub}
\end{figure}

\vspace{2ex}
\noindent
\textbf{More visualization on predicted flow fields:} We also give more flow visualization examples in Fig.~\ref{fig:flow_map_sub}. The flow color encoding is shown below. The left part is the colormap, while the right part is the direction map.

\begin{figure}[t]
	\centering
	\includegraphics[width=0.85\linewidth]{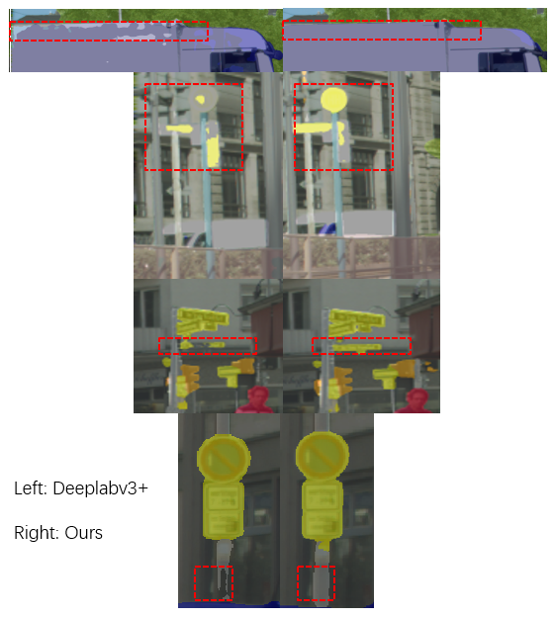}
	\vspace{-5mm}
	\caption{
	   Mask prediction examples based on Deeplabv3+. The refined parts are shown in red boxes.
	}
	\label{fig:predict_mask}
\end{figure}

\vspace{2ex}
\noindent
\textbf{More predicted fine-grained mask visualization:}
Fig.~\ref{fig:predict_mask} gives more fine-grained mask prediction which are shown in the red boxes.

	\clearpage

	\bibliographystyle{splncs}
	\bibliography{egbib}
\end{document}